\documentclass{bmvc2k}

\title{CR-JEPA: Cross-Modal Joint-Embedding Predictive Learning for Remote Sensing Image Retrieval}

\addauthor{Md Aminur Hossain}{md.aminurhossain@gmail.com}{1}
\addauthor{Ayush V. Patel}{ayu020503@gmail.com}{2}
\addauthor{Nitant Dube}{nitant@sac.isro.gov.in}{1}
\addauthor{Biplab Banerjee}{getbiplab@gmail.com}{2}

\addinstitution{
 Space Applications Centre,\\
 Indian Space Research Organisation, \\
 Ahmedabad, India
}
\addinstitution{
 Centre of Studies in Resources Engineering,\\
 Indian Institute of Technology Bombay,\\
 Mumbai, India
}

\runninghead{Hossain et al.}{CR-JEPA: Cross-Modal Remote Sensing Image Retrieval}

\usepackage{amsmath}
\usepackage{amssymb}
\usepackage{booktabs}
\usepackage{enumitem}
\usepackage{graphicx}
\usepackage{multirow}
\usepackage{xspace}

\newcommand{\method}{CR-JEPA\xspace}

\begin{document}

\maketitle

\begin{abstract}
Cross-modal remote sensing image retrieval aims to retrieve semantically related scenes across heterogeneous sensing modalities. This remains challenging because paired observations may differ substantially in imaging physics, spatial resolution, spectral configuration, and visual appearance. Moreover, a single retrieval projection trained with one objective may be insufficient to jointly support cross-modal semantic alignment and same-modal neighbourhood preservation. We propose CR-JEPA, a Cross-modal Retrieval Joint-Embedding Predictive Architecture for dual-modality remote sensing retrieval. The model uses modality-specific stems, a shared transformer trunk, and JEPA-style predictive objectives to estimate masked latent target features within and across modalities. Inspired by LeJEPA, we apply Sketched Isotropic Gaussian Regularization to raw retrieval projections to stabilize embeddings and mitigate collapse. CR-JEPA further employs a decoupled-head design with a unified retrieval head for same-modal retrieval and a cross-modal retrieval head for cross-modal search. We evaluate CR-JEPA on BEN-14K, CBRSIR\_VS, and DSRSID. On BEN-14K, CR-JEPA improves S1$\rightarrow$S2 retrieval from 61.23\% to 75.82\% and S2$\rightarrow$S1 retrieval from 63.73\% to 75.40\% over X-JEPA, while also achieving competitive same-modal retrieval with fewer parameters.
\end{abstract}

\section{Introduction}
\label{sec:introduction}

The scale of Earth observation (EO) archives has grown rapidly, leading to a greater need for efficient retrieval systems that can find semantically related scenes within large collections~\cite{zhou2023rsir,tong2022survey}. Remote sensing image retrieval (RSIR) allows for exploring archives, performing case-based analysis, transferring weak annotations, monitoring, and searching when a fixed classifier is not available or is too limiting~\cite{li2021rsbigsurvey,zhou2023rsir}. Relying on metadata-based indexing alone often does not work well. Remote sensing archives include various sensor types, complex scene semantics, large variations within classes, and multi-label land-cover content~\cite{tong2022survey,chaudhuri2018multilabelgraph}.


Cross-modal remote sensing retrieval is more challenging than conventional unimodal retrieval, since the query and gallery images may come from different sensing modalities~\cite{sumbul2022sscmir,choudhury2026xjepa}. Often these modalities describe the same scene by different physical processes. Optical imagery captures spectral reflectance, colour and visual texture, while synthetic aperture radar (SAR) imagery captures surface structure, roughness and backscatter properties~\cite{fuller2023croma,sumbul2022sscmir}. The panchromatic and multispectral images also have different spatial and spectral resolution, which causes another modality gap in cross-source retrieval~\cite{li2018sidhcnn}. Thus, paired observations may be geographically or semantically related but visually very different. This modality gap means direct feature comparison is unreliable and needs representations that can align cross-modal semantics while keeping useful same-modal neighbourhood structure~\cite{choudhury2026xjepa,hackstein2025csmae}.

Existing self-supervised approaches have yet to fully resolve this problem. Contrastive cross-modal methods successfully align paired views, but their performance can be sensitive to false negatives, batch composition, and the construction of reliable positive pairs~\cite{sumbul2022sscmir}. As a scalable alternative, masked modeling bypasses some of these issues, but often at the cost of wasting model capacity on reconstructing low-level, modality-specific signals. Since retrieval relies heavily on semantic similarity rather than pixel-level fidelity, this reconstruction bottleneck is far from ideal~\cite{hackstein2025csmae,fuller2023croma}. Joint-Embedding Predictive Architectures (JEPAs) elegantly sidestep this issue by predicting latent target representations instead of raw pixels~\cite{assran2023ijepa}. Recent remote sensing variants, such as REJEPA and X-JEPA, highlight the effectiveness of feature-space prediction for retrieval~\cite{choudhury2025rejepa,choudhury2026xjepa}. However, there is still significant room to improve how these frameworks learn across heterogeneous modality pairs. Specifically, existing models struggle when required to support both cross-modal semantic alignment and same-modal neighbourhood preservation at the same time.

To address this limitation, we propose \method, a Cross-modal Retrieval Joint-Embedding Predictive Architecture for heterogeneous dual-modality remote sensing retrieval. CR-JEPA combines modality-specific stems, a shared transformer trunk, same-modal and cross-modal latent predictive learning, and SIGReg-based Gaussian embedding regularization inspired by LeJEPA~\cite{balestriero2025lejepa}. The model further uses a compact decoupled-head design with a unified retrieval head for same-modal retrieval and a cross-modal retrieval head for cross-modal search. The main contributions of this work are as follows:
\begin{itemize}[leftmargin=1.5em]
    \item We propose \method, a cross-modal joint-embedding predictive learning framework for heterogeneous dual-modality remote sensing image retrieval.

    \item We design a modality-adaptive architecture that combines modality-specific stems, a shared transformer trunk, and a compact decoupled-head design to separate low-level sensor adaptation, high-level semantic reasoning, and retrieval-space learning.

    \item We formulate same-modal and cross-modal latent predictive objectives together with SIGReg-based Gaussian embedding regularization to improve representation stability, mitigate collapse, and learn semantically aligned retrieval embeddings.

    \item We evaluate \method on three dual-modality benchmarks: BEN-14K, CBRSIR\_VS, and DSRSID, covering Sentinel-1/Sentinel-2, optical/SAR, and panchromatic/multispectral retrieval settings under both same-modal and cross-modal retrieval protocols.
\end{itemize}

\section{Related Work}
\label{sec:related}

\subsection{Remote Sensing Image Retrieval}

Remote sensing image retrieval (RSIR) has progressed from handcrafted descriptors and graph-based matching to deep representation learning systems. Recent surveys show that deep models have substantially improved retrieval from large-scale remote sensing archives~\cite{sudha2019review,li2021rsbigsurvey,tong2022survey}. However, RSIR remains challenging because remote sensing images contain diverse sensor characteristics, complex scene semantics, large intra-class variation, and multi-label land-cover content~\cite{zhou2023rsir}. Multi-label retrieval is especially important in Earth observation, since a single image patch often contains multiple land-cover categories. Prior studies have explored semi-supervised graph-based retrieval, dense labels, and fully convolutional representations for multi-label RSIR~\cite{chaudhuri2018multilabelgraph,shao2018dlrsd,shao2020fcnrsir}. Protocol-oriented studies further show that single-label benchmarks can become saturated and that multi-label retrieval requires suitable ranking metrics and evaluation protocols~\cite{imbriaco2022multilabel}.

\subsection{Cross-Modal Remote Sensing Retrieval}

Cross-modal remote sensing retrieval aims to retrieve semantically related scenes across heterogeneous sensing modalities. This setting includes modality pairs such as optical/SAR and panchromatic/multispectral, where low-level appearance may differ substantially even when the scene semantics are similar. Early supervised approaches, such as CMIR-Net, demonstrated the feasibility of cross-modal remote sensing retrieval, but relied on labeled paired data to learn semantic alignment~\cite{chaudhuri2020cmirnet}. Source-invariant hashing methods, such as SIDHCNN, further addressed cross-source retrieval involving high-resolution optical, panchromatic, and multispectral imagery~\cite{li2018sidhcnn}. Semantic-preserving hashing methods, such as MsEspH, extended this direction to multisensor optical/SAR retrieval with explicit semantic preservation~\cite{sun2022msesph}. Knowledge-distillation-based cross-source retrieval has also been explored to improve modality-invariant representation learning, including discriminative distillation networks and ensemble distillation frameworks~\cite{xiong2020discriminative,ma2021crosssource}.

Beyond hashing-based methods, several approaches have explored stronger cross-modal alignment mechanisms. Unified attention networks have been used to enhance modality interaction for remote sensing retrieval~\cite{choudhury2024uan}, while hypergraph neural networks model higher-order relationships among multimodal samples~\cite{yu2023hgnlsf}. Robust correlation learning and contrastive prototype alignment have also been studied to reduce modality discrepancy between query and gallery representations~\cite{wang2024hac,h2026xclpa}. Recent correlation-aware contrastive methods further exploit spatiotemporal context for cross-modal remote sensing retrieval~\cite{zhu2024ccls2t}. Recent work has also investigated frequency-guided distillation and generative pre-alignment for lightweight or cross-domain remote sensing retrieval~\cite{xu2026frequency,huang2025otpfcnet}. 
SS-CMIR uses contrastive learning to align cross-modal representations without dense manual annotations~\cite{sumbul2022sscmir}. Recent multimodal pretraining frameworks further broaden Earth observation representation learning, including CROMA~\cite{fuller2023croma}, DeCUR~\cite{wang2024decur}, and large-scale foundation models such as SkySense, AnySat, and CSMoE~\cite{guo2024skysense,astruc2025anysat,hackel2025csmoe}. Large-scale multimodal datasets and semantically grounded pretraining resources, including SSL4EO-S12 and GeoMeld, also support scalable multimodal learning for Earth observation~\cite{wang2023ssl4eo,hasan2026geomeld}.

\subsection{Self-Supervised Retrieval and Predictive Representation Learning}

Masked image modeling has become a strong paradigm for remote sensing representation learning. SatMAE, ScaleMAE, and SatMAE++ adapt masked autoencoding to temporal, multispectral, and multi-scale satellite imagery~\cite{cong2022satmae,reed2023scalemae,noman2024satmaepp}. CSMAE is particularly relevant to retrieval because it studies masked autoencoding for sensor-agnostic image retrieval under both unimodal and cross-modal settings~\cite{hackstein2025csmae}. However, reconstruction-based objectives primarily optimize pixel or signal recovery, which can emphasize modality-specific low-level details rather than modality-common semantics needed for retrieval.

Joint-Embedding Predictive Architectures (JEPAs) provide an alternative by predicting target representations directly in latent space instead of reconstructing pixels~\cite{assran2023ijepa}. Recent studies also highlight the shift from pure alignment objectives toward prediction-based self-supervised learning~\cite{dutta2026alignmentprediction}. In remote sensing, REJEPA applies JEPA principles to unimodal image retrieval~\cite{choudhury2025rejepa}, while X-JEPA extends predictive learning to cross-modal retrieval using target-modality embedding prediction and prediction-space alignment~\cite{choudhury2026xjepa}. Building on these works, \method combines same-modal and cross-modal latent prediction with SIGReg regularization and decoupled retrieval heads for same-modal and cross-modal retrieval.

\subsection{Latent-Euclidean JEPA and SIGReg}

A central challenge in joint-embedding predictive learning is avoiding collapse while maintaining useful embedding geometry. Existing self-supervised methods use negative samples, stop-gradient learning, teacher-student networks, whitening, or variance-covariance regularization to prevent degenerate solutions~\cite{bardes2021vicreg,assran2023ijepa}. LeJEPA provides a Latent-Euclidean perspective by encouraging embeddings to follow an isotropic Gaussian distribution and introduces Sketched Isotropic Gaussian Regularization (SIGReg) as an efficient projection-based distribution matching objective~\cite{balestriero2025lejepa}. Inspired by this idea, CR-JEPA applies SIGReg to raw retrieval projections before $\ell_2$ normalization, encouraging stable retrieval embeddings for heterogeneous dual-modality remote sensing retrieval.

\section{Proposed Methodology}
\label{sec:method}

\subsection{Problem Formulation}

We consider a paired dual-modality remote sensing retrieval setting. Let
\begin{equation}
\mathcal{D}=\{(x_i^{(a)},x_i^{(b)},y_i)\}_{i=1}^{N}
\end{equation}
denote a dataset of $N$ paired observations, where $x_i^{(a)}$ and $x_i^{(b)}$ are two modality views of the same scene or a semantically corresponding sample, and $y_i$ denotes the semantic label annotation. The two modalities depend on the dataset: BEN-14K uses Sentinel-1 and Sentinel-2 imagery, CBRSIR\_VS uses optical and SAR imagery, and DSRSID uses panchromatic and multispectral imagery. Thus, the proposed formulation is not restricted to a single sensor pair, but applies to heterogeneous dual-modality remote sensing retrieval.

The retrieval task is evaluated in four directions:
\begin{equation}
a\rightarrow a,\qquad b\rightarrow b,\qquad a\rightarrow b,\qquad b\rightarrow a .
\end{equation}

The first two relate to same-modal retrieval, while the last two relate to cross-modal retrieval. For BEN-14K, relevance is based on the overlap of multiple labels between the query and gallery samples. For CBRSIR\_VS and DSRSID, where each image pair fits into one semantic class, relevance is based on single-label class equality. The goal is to learn embeddings that support both cross-modal semantic alignment and same-modal neighbourhood preservation.

\subsection{Architecture Overview}

\begin{figure}[ht]
    \centering
    \includegraphics[width=\linewidth]{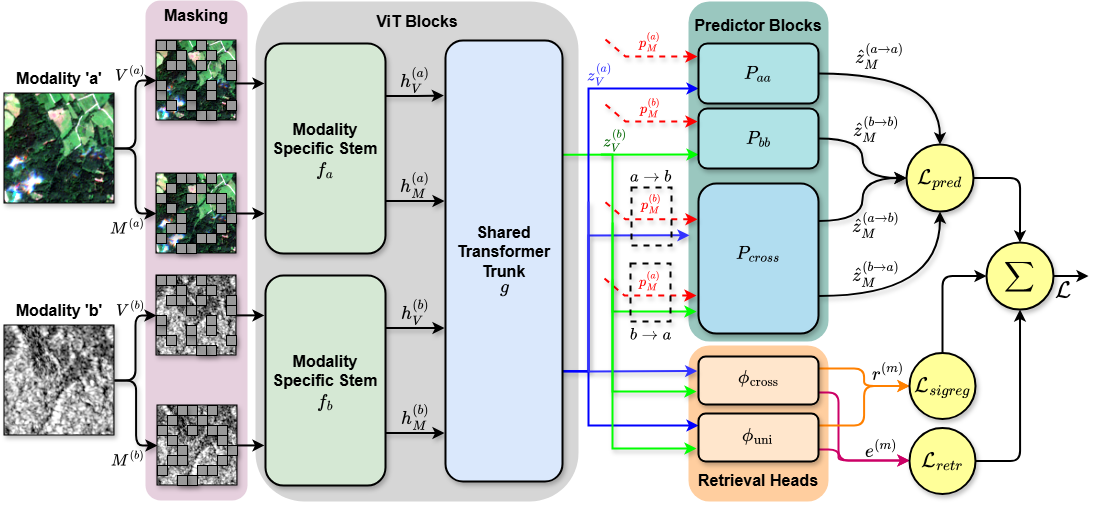}
    \caption{Overview of \method. Each modality is first processed by a modality-specific stem and then passed through a shared transformer trunk. Same-modal and cross-modal predictive branches learn masked latent target prediction in feature space. A unified retrieval head supports same-modal retrieval, while a cross-modal retrieval head supports cross-modal search. SIGReg regularizes the raw retrieval projections.}
    \label{fig:architecture}
\end{figure}

\method is a cross-modal retrieval joint-embedding predictive learning framework. As illustrated in Figure~\ref{fig:architecture}, it is built around three design principles: modality-specific adaptation, shared semantic reasoning, and retrieval-space specialization. Given an input pair $(x^{(a)},x^{(b)})$, each modality is first processed by its own stem to handle differences in channel configuration, spatial resolution, and low-level sensor statistics. The resulting token sequences are then passed through a shared transformer trunk, which learns modality-common semantic representations. Predictive heads learn latent feature forecasting within and across modalities, while retrieval heads learn compact embedding spaces for alignment and retrieval.

Let $f_a$ and $f_b$ denote the modality-specific stems, and let $g$ denote the shared transformer trunk. For a modality $m\in\{a,b\}$, the stem converts the input into a sequence of patch tokens:
\begin{equation}
h^{(m)} = f_m(x^{(m)}),
\end{equation}
where $h^{(m)}\in\mathbb{R}^{T\times D}$, $T$ is the number of tokens, and $D$ is the embedding dimension. The shared trunk maps the modality-specific tokens into a common semantic processing space:
\begin{equation}
z^{(m)} = g(h^{(m)}).
\end{equation}
Unlike a fully shared encoder, the modality-specific stems allow the model to adapt to sensor-dependent low-level characteristics. Unlike fully separate encoders, the shared trunk encourages semantic coupling across modalities.

\subsection{Modality-Specific Stems and Shared Semantic Trunk}

Remote sensing modalities can differ in channel number, spatial resolution, radiometric properties, and imaging physics. For example, SAR and optical images represent different physical responses, while panchromatic and multispectral images differ in spatial and spectral resolution. To handle these differences, \method uses a separate stem for each modality. Each stem consists of a patch embedding layer followed by modality-specific positional embeddings. This allows the model to map heterogeneous inputs into a common token dimension while preserving modality-dependent low-level structure.

After stem tokenization, token sequences from both modalities are processed by the same transformer trunk. The shared trunk encourages learning modality-common semantic representations instead of maintaining two independent feature spaces. This design provides a balance between sensor adaptation and semantic sharing: the stems absorb low-level modality differences, while the trunk learns higher-level scene structure useful for retrieval.

\subsection{Same-Modal and Cross-Modal Predictive Learning}

\begin{figure}[ht]
    \centering
    \includegraphics[width=\linewidth]{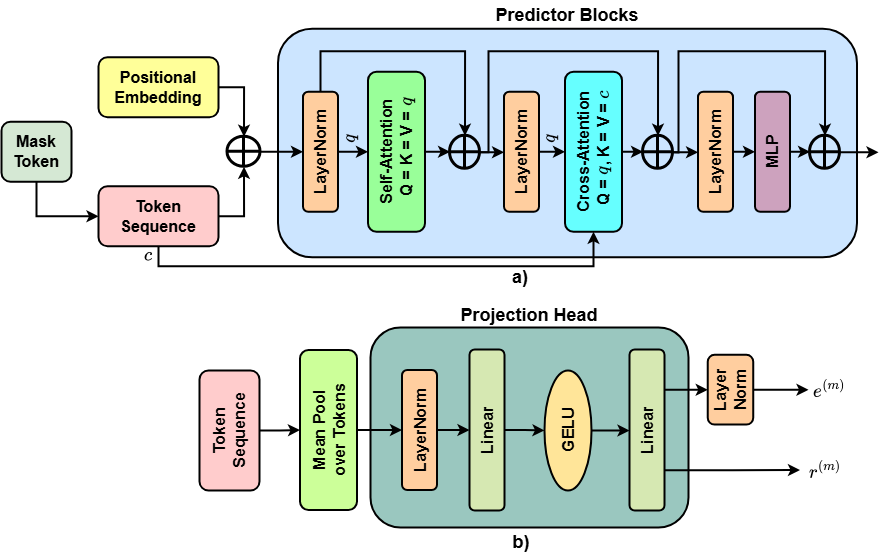}
    \caption{Predictive and retrieval components of \method. (a) Query-based predictor for masked latent target prediction; (b) retrieval head producing raw projections for SIGReg and normalized embeddings for retrieval.}
    \label{fig:pred_retr}
\end{figure}

The predictive objective follows a JEPA-style masked-token protocol in latent space, as illustrated in Figure~\ref{fig:pred_retr}(a). For each modality $m\in\{a,b\}$, token indices are split into visible indices $V^{(m)}$ and masked indices $M^{(m)}$ using a random masking strategy with a fixed mask ratio. The visible tokens are used as context, while the masked tokens provide latent prediction targets:
\begin{equation}
z_{V}^{(m)} = g\!\left(f_m(x^{(m)};V^{(m)})\right), \qquad
z_{M}^{(m)} = g\!\left(f_m(x^{(m)};M^{(m)})\right),
\end{equation}
where $f_m(x^{(m)};I)$ denotes stem tokenization followed by selection of the token subset indexed by $I$. The model predicts masked target tokens directly in feature space; no pixel reconstruction branch is used.

The model includes two same-modal predictors and one shared cross-modal predictor. The same-modal predictors estimate masked target tokens within the same modality:
\begin{equation}
\hat{z}^{(a\rightarrow a)}_{M}=P_{aa}(z_{V}^{(a)},p_{M}^{(a)}), \qquad
\hat{z}^{(b\rightarrow b)}_{M}=P_{bb}(z_{V}^{(b)},p_{M}^{(b)}),
\end{equation}
where $p_{M}^{(m)}$ denotes the positional embeddings of the masked target locations for modality $m$. The shared cross-modal predictor estimates masked target tokens of one modality from the visible context tokens of the other modality:
\begin{equation}
\hat{z}^{(a\rightarrow b)}_{M}=P_{\mathrm{cross}}(z_{V}^{(a)},p_{M}^{(b)}), \qquad
\hat{z}^{(b\rightarrow a)}_{M}=P_{\mathrm{cross}}(z_{V}^{(b)},p_{M}^{(a)}).
\end{equation}
Each predictor is implemented using learnable mask queries, target-position embeddings, self-attention, cross-attention to the visible context tokens, and an MLP block. The same-modal predictors encourage within-modality feature forecasting, while the cross-modal predictor encourages semantic prediction across modalities.

The predictive loss is defined as
\begin{align}
\mathcal{L}_{\mathrm{pred}} =
&\lambda_{aa}\left\|\hat{z}^{(a\rightarrow a)}_{M}-z^{(a)}_{M}\right\|_2^2
+\lambda_{bb}\left\|\hat{z}^{(b\rightarrow b)}_{M}-z^{(b)}_{M}\right\|_2^2 \nonumber\\
&+\lambda_{ab}\left\|\hat{z}^{(a\rightarrow b)}_{M}-z^{(b)}_{M}\right\|_2^2
+\lambda_{ba}\left\|\hat{z}^{(b\rightarrow a)}_{M}-z^{(a)}_{M}\right\|_2^2 .
\end{align}
This objective encourages the model to learn both within-modality predictive structure and cross-modal semantic correspondence without reconstructing raw sensor measurements.

\subsection{Decoupled Retrieval Heads}

A central design choice in \method is to avoid relying on a single retrieval projection for all retrieval behaviours. Same-modal retrieval requires preserving class-consistent neighbourhoods within each modality, whereas cross-modal retrieval requires aligning semantically corresponding samples across heterogeneous modalities. As illustrated in Figure~\ref{fig:pred_retr}(b), we therefore use a compact decoupled-head design with two complementary retrieval heads: i) a unified retrieval head $\phi_{\mathrm{uni}}$ for same-modal retrieval, and ii) a cross-modal retrieval head $\phi_{\mathrm{cross}}$ for cross-modal retrieval.

The visible token sequence is mean-pooled before projection:
\begin{equation}
\bar{z}^{(m)}=\frac{1}{|V^{(m)}|}\sum_{t\in V^{(m)}} z^{(m)}_{V,t},
\end{equation}
where $m\in\{a,b\}$ denotes the modality. Each retrieval head produces a raw projection $r$ and an $\ell_2$-normalized embedding $e$:
\begin{equation}
r_{\mathrm{uni}}^{(m)},e_{\mathrm{uni}}^{(m)}
=\phi_{\mathrm{uni}}(\bar{z}^{(m)}),
\end{equation}
\begin{equation}
r_{\mathrm{cross}}^{(m)},e_{\mathrm{cross}}^{(m)}
=\phi_{\mathrm{cross}}(\bar{z}^{(m)}).
\end{equation}

The unified head is used for same-modal retrieval, where query and gallery samples belong to the same modality. The cross-modal retrieval head is used for cross-modal search, where query and gallery samples come from different modalities. This separation allows \method to preserve within-modality neighbourhood structure while learning a dedicated cross-modal embedding space for heterogeneous retrieval.

\subsection{SIGReg-Based Embedding Regularization}

To stabilize the learned retrieval embeddings, \method incorporates Sketched Isotropic Gaussian Regularization (SIGReg), inspired by LeJEPA~\cite{balestriero2025lejepa}. SIGReg encourages the distribution of learned embeddings to approach an isotropic Gaussian distribution, helping mitigate representation collapse and improve latent-space conditioning. We apply SIGReg to the raw pre-normalized retrieval projections from the cross-modal retrieval head and the unified retrieval head:
\begin{equation}
r_{\mathrm{cross}}^{(a)},\quad
r_{\mathrm{cross}}^{(b)},\quad
r_{\mathrm{uni}}^{(a)},\quad
r_{\mathrm{uni}}^{(b)}.
\end{equation}

Given a mini-batch of raw projections $R\in\mathbb{R}^{B\times D}$, SIGReg samples $J$ random unit projection directions $\{u_j\}_{j=1}^{J}$ and computes one-dimensional projections
\begin{equation}
s_{ij}=R_i^\top u_j .
\end{equation}
It then matches the empirical characteristic function of these projected samples to the characteristic function of a standard Gaussian,
\begin{equation}
\phi(t_k)=\exp(-t_k^2/2),
\end{equation}
over frequency points $\{t_k\}_{k=1}^{K}$. The loss is
\begin{equation}
\mathcal{L}_{\mathrm{sigreg}}(R)
=
\frac{B}{J}
\sum_{j=1}^{J}
\sum_{k=1}^{K}
\omega_k
\left[
\left(
\frac{1}{B}\sum_{i=1}^{B}\cos(t_k s_{ij})-\phi(t_k)
\right)^2
+
\left(
\frac{1}{B}\sum_{i=1}^{B}\sin(t_k s_{ij})
\right)^2
\right],
\end{equation}
where $\omega_k$ denotes the numerical integration weight. In our implementation, $t_k$ is uniformly sampled in $[0,3]$, and $\omega_k$ follows a trapezoidal quadrature weight multiplied by $\phi(t_k)$. The final SIGReg term is applied to the raw retrieval projections as
\begin{equation}
\mathcal{L}_{\mathrm{sigreg}}
=
\frac{1}{4}
\Big[
\mathcal{L}_{\mathrm{sigreg}}(r_{\mathrm{cross}}^{(a)})
+
\mathcal{L}_{\mathrm{sigreg}}(r_{\mathrm{cross}}^{(b)})
+
\mathcal{L}_{\mathrm{sigreg}}(r_{\mathrm{uni}}^{(a)})
+
\mathcal{L}_{\mathrm{sigreg}}(r_{\mathrm{uni}}^{(b)})
\Big].
\end{equation}
This regularization complements the predictive and retrieval losses by encouraging a stable, non-degenerate distribution of retrieval embeddings.

\subsection{Training Objective}

The training objective combines latent predictive learning, cross-modal retrieval learning, unified-space alignment, and SIGReg-based embedding regularization. Let $\mathrm{NCE}(Q,K)$ denote a symmetric batchwise InfoNCE loss between two sets of normalized embeddings.

The cross-modal retrieval loss is applied to the cross-modal retrieval head:
\begin{equation}
\mathcal{L}_{\mathrm{cross}}
=
\mathrm{NCE}(e_{\mathrm{cross}}^{(a)},e_{\mathrm{cross}}^{(b)}).
\end{equation}
This loss encourages paired observations from the two modalities to become semantically aligned in the cross-modal retrieval space used for $a\rightarrow b$ and $b\rightarrow a$ search.

The unified-space alignment loss is applied to the unified head, which is used for same-modal retrieval at inference time. It combines cross-modal InfoNCE with direct cosine alignment:
\begin{equation}
\mathcal{L}_{\mathrm{uni}}
=
\mathrm{NCE}(e_{\mathrm{uni}}^{(a)},e_{\mathrm{uni}}^{(b)})
+
\frac{1}{B}\sum_{i=1}^{B}
\left(
1-\cos(e_{\mathrm{uni},i}^{(a)},e_{\mathrm{uni},i}^{(b)})
\right),
\end{equation}
where $B$ is the mini-batch size. Although this loss is computed using paired cross-modal samples, the resulting unified embeddings are used for same-modal retrieval, where they help preserve semantically consistent neighbourhoods within each modality.

The weighted retrieval objective is therefore
\begin{equation}
\mathcal{L}_{\mathrm{retr}}
=
\lambda_{\mathrm{cross}}\mathcal{L}_{\mathrm{cross}}
+
\lambda_{\mathrm{uni}}\mathcal{L}_{\mathrm{uni}}.
\end{equation}

The final training objective is
\begin{equation}
\mathcal{L}
=
\mathcal{L}_{\mathrm{pred}}
+
\mathcal{L}_{\mathrm{retr}}
+
\lambda_{\mathrm{sigreg}}\mathcal{L}_{\mathrm{sigreg}}.
\end{equation}
This objective jointly optimizes feature-space prediction, cross-modal retrieval alignment, unified embedding learning, and embedding distribution regularity.

\subsection{Inference Protocol}

At inference time, each image is encoded by its corresponding modality-specific stem and the shared transformer trunk. The pooled representation is then projected using the retrieval head associated with the retrieval direction.

For same-modal retrieval, \method uses the unified retrieval embedding:
\begin{equation}
a\rightarrow a,\qquad b\rightarrow b
\quad \text{use} \quad e_{\mathrm{uni}}.
\end{equation}
For cross-modal retrieval, \method uses the cross-modal retrieval embedding:
\begin{equation}
a\rightarrow b,\qquad b\rightarrow a
\quad \text{use} \quad e_{\mathrm{cross}}.
\end{equation}
Nearest-neighbour retrieval is performed using cosine similarity between normalized embeddings. For same-modal retrieval, trivial self-matches are excluded when the query and gallery contain the same indexed sample.

\section{Experimental Setup}
\label{sec:experiments}

\subsection{Datasets}

We evaluate \method on three dual-modality remote sensing retrieval benchmarks covering Sentinel-1/Sentinel-2, optical/SAR, and panchromatic/multispectral retrieval. Detailed dataset descriptions are provided in the supplementary material.

\textbf{BEN-14K.}
BEN-14K is derived from BigEarthNet-MM and contains paired Sentinel-1 SAR and Sentinel-2 multispectral observations with multi-label land-cover annotations~\cite{sumbul2021bigearthnetmm,bigearthnet}. Following prior sensor-agnostic retrieval protocols, we use BEN-14K to evaluate same-modal and cross-modal retrieval under multi-label relevance~\cite{hackstein2025csmae,choudhury2026xjepa}.

\textbf{CBRSIR\_VS.}
CBRSIR\_VS is an optical/SAR dual-modality dataset for cross-modal remote sensing retrieval~\cite{sun2022msesph}. It contains 26,901 paired RGB optical and SAR images from 10 semantic classes. In our protocol, the two modalities are RGB optical and SAR.

\textbf{DSRSID.}
DSRSID is a dual-source panchromatic/multispectral dataset for cross-source remote sensing retrieval~\cite{li2018sidhcnn}. It contains 80,000 paired samples from eight scene classes. In our protocol, the two modalities are panchromatic and multispectral images.

\subsection{Evaluation Metrics}

We evaluate retrieval in four directions: $a\rightarrow a$, $b\rightarrow b$, $a\rightarrow b$, and $b\rightarrow a$, covering both same-modal and cross-modal retrieval. Trivial self-matches are excluded in same-modal retrieval. For BEN-14K, relevance is based on multi-label overlap, and we report F1@5 following prior protocols~\cite{hackstein2025csmae,choudhury2026xjepa}. For CBRSIR\_VS and DSRSID, relevance is based on single-label class equality, and we report global mAP and P@5. Detailed metric definitions are provided in the supplementary material.

\subsection{Implementation Details}

All images are resized to $224\times224$ before tokenization. We use a patch size of $16$, embedding dimension $512$, $8$ attention heads, predictor depth $6$, retrieval dimension $256$, mask ratio $0.5$, and shared transformer trunk depth $12$. Each modality is processed by its own modality-specific stem, allowing the model to handle different channel configurations across datasets. For BEN-14K, Sentinel-1 uses two channels and Sentinel-2 uses 12 channels. For CBRSIR\_VS, the optical branch uses RGB images and the SAR branch uses the SAR intensity image. For DSRSID, the panchromatic branch uses one channel and the multispectral branch uses four channels.

Training uses AdamW~\cite{loshchilov2017adamw} with learning rates scheduled from $10^{-4}$ to $10^{-3}$ and then to $10^{-6}$, weight decay $0.04$, gradient clipping at $1.0$, cosine scheduling with $15$ warmup epochs, and automatic mixed precision. Unless otherwise stated, models are trained for $400$ epochs with batch size $512$ and evaluated with batch size $256$ on NVIDIA A100 80\,GB GPUs. The full objective combines same-modal and cross-modal predictive losses, cross-modal retrieval loss, unified retrieval loss, and SIGReg regularization.

\subsection{Baselines}

We compare \method with representative self-supervised, multimodal, and JEPA-based remote sensing retrieval baselines. For BEN-14K, we use published benchmark results when the evaluation protocol is consistent with our setting. For CBRSIR\_VS and DSRSID, we evaluate adapted REJEPA and X-JEPA implementations under the same preprocessing, training, and evaluation protocol as \method.

For BEN-14K, we compare with MAE~\cite{he2022mae}, MAE-RVSA, SatMAE~\cite{cong2022satmae}, SatMAE++~\cite{noman2024satmaepp}, ScaleMAE~\cite{reed2023scalemae}, CrossMAE~\cite{tang2023crossscale}, CSMAE-SESD~\cite{hackstein2025csmae}, SkySense~\cite{guo2024skysense}, CROMA~\cite{fuller2023croma}, DeCUR~\cite{wang2024decur}, SS-CMIR~\cite{sumbul2022sscmir}, REJEPA~\cite{choudhury2025rejepa}, and X-JEPA~\cite{choudhury2026xjepa}. This benchmark enables direct comparison with recent JEPA-style retrieval methods because REJEPA and X-JEPA report results on BEN-14K.

For CBRSIR\_VS and DSRSID, publicly reported results from recent self-supervised retrieval methods are not available under a consistent evaluation protocol. Therefore, we evaluate REJEPA~\cite{choudhury2025rejepa} and X-JEPA~\cite{choudhury2026xjepa} using adapted implementations under the same preprocessing, training, and evaluation protocol as \method. Since the original REJEPA and X-JEPA papers do not report results on these datasets, we use the same splits, input resolution, retrieval directions, and evaluation metrics for a controlled comparison.

\section{Results and Analysis}
\label{sec:results}

\subsection{Main Results on BEN-14K}

Table~\ref{tab:main_results_ben14k} reports the main F1@5 comparison on BEN-14K. The published baseline values follow the benchmark layout used in X-JEPA, allowing \method to be compared under the same four-direction retrieval protocol. The results show that \method is particularly effective in cross-modal retrieval. Compared with X-JEPA, \method improves S1$\rightarrow$S2 from 61.23 to 75.82 and S2$\rightarrow$S1 from 63.73 to 75.40. This corresponds to absolute gains of 14.59 and 11.67 F1@5 points, respectively, establishing the best performance among the compared methods in cross-modal retrieval on BEN-14K.

In same-modal retrieval, \method also achieves strong performance. For S1$\rightarrow$S1 retrieval, it improves over X-JEPA from 72.98 to 75.11, yielding a gain of 2.13 F1@5 points. In S2$\rightarrow$S2 retrieval, \method attains 82.87, slightly surpassing X-JEPA (82.65) and achieving the best overall performance among all compared methods. Overall, these results indicate that \method not only significantly enhances cross-modal semantic alignment but also maintains, and in some cases improves, same-modal neighbourhood preservation within a more parameter-efficient architecture.

\begin{table*}[ht]
\begin{center}
\small
\resizebox{\textwidth}{!}{%
\begin{tabular}{lccccc}
\toprule
\multirow{2}{*}{Method} & \multirow{2}{*}{Params (M)} & \multicolumn{2}{c}{Same-modal F1@5 (\%)} & \multicolumn{2}{c}{Cross-modal F1@5 (\%)} \\
\cmidrule(lr){3-4} \cmidrule(lr){5-6}
 &  & S1$\rightarrow$S1 & S2$\rightarrow$S2 & S1$\rightarrow$S2 & S2$\rightarrow$S1 \\
\midrule
MAE & 224.87 & 60.81 & 72.04 & 41.78 & 46.12 \\
MAE-RVSA & 227.75 & 55.40 & 71.47 & 36.66 & 38.05 \\
SatMAE & 329.40 & 70.86 & 78.71 & 49.57 & 52.48 \\
SatMAE++ & 329.14 & 67.29 & 76.48 & 50.21 & 54.98 \\
ScaleMAE & 284.35 & 62.73 & -- & -- & -- \\
CrossMAE & 250.57 & 66.45 & 71.28 & 49.46 & 48.71 \\
CSMAE-SESD (Disjoint) & 210.64 & 70.62 & 39.01 & 38.74 & 38.42 \\
SkySense & 398.04 & 69.87 & 73.42 & 50.26 & 52.11 \\
CROMA & 310.54 & 68.48 & 72.71 & 46.53 & 48.61 \\
DeCUR & 250.54 & 71.26 & 75.36 & 40.78 & 41.83 \\
REJEPA & 197.09 & \textbf{76.38} & 75.42 & 55.46 & 56.32 \\
X-JEPA & 172.86 & 72.98 & \underline{82.65} & \underline{61.23} & \underline{63.73} \\
\method & 117.93 & \underline{75.11} & \textbf{82.87} & \textbf{75.82} & \textbf{75.40} \\
\bottomrule
\end{tabular}
}
\end{center}
\caption{Main BEN-14K retrieval results following the X-JEPA benchmark protocol. The reported metric is F1@5. Best and second-best results are shown in bold and underlined, respectively.}
\label{tab:main_results_ben14k}
\end{table*}

\paragraph{Computational Cost.}
On BEN-14K, \method requires approximately 9.6 GFLOPs per image with an average inference time of 17 ms, compared with X-JEPA at 10.8 GFLOPs and 20 ms. Inference time is reported as average per-image latency measured on an NVIDIA A100 80\,GB GPU using $224\times224$ inputs, batch size $256$, and automatic mixed precision. The shared transformer trunk and lightweight predictive heads enable improved cross-modal alignment without adding substantial computational overhead, showing that \method improves retrieval accuracy while remaining computationally efficient.

\subsection{Main Results on CBRSIR\_VS and DSRSID}

Table~\ref{tab:cbrsir_dsrsid_results} reports retrieval performance under single-label relevance on two cross-modal remote sensing benchmarks. For CBRSIR\_VS, we evaluate RGB$\rightarrow$RGB, SAR$\rightarrow$SAR, RGB$\rightarrow$ SAR, and SAR$\rightarrow$RGB retrieval. For DSRSID, we evaluate PAN$\rightarrow$PAN, MS$\rightarrow$MS, PAN$\rightarrow$MS, and MS$\rightarrow$PAN retrieval. We report global mAP and P@5.

\begin{table*}[t]
\centering
\small
\setlength{\tabcolsep}{3pt}
\renewcommand{\arraystretch}{1.15}
\resizebox{\textwidth}{!}{%
\begin{tabular}{llcccccccc}
\toprule
Dataset & Method
& \multicolumn{2}{c}{RGB$\rightarrow$RGB}
& \multicolumn{2}{c}{SAR$\rightarrow$SAR}
& \multicolumn{2}{c}{RGB$\rightarrow$SAR}
& \multicolumn{2}{c}{SAR$\rightarrow$RGB} \\
\cmidrule(lr){3-4}\cmidrule(lr){5-6}\cmidrule(lr){7-8}\cmidrule(lr){9-10}
& & mAP & P@5 & mAP & P@5 & mAP & P@5 & mAP & P@5 \\
\midrule
\multirow{3}{*}{CBRSIR\_VS}
& REJEPA  
& 75.74 & 81.45 
& 63.21 & 67.92 
& 66.88 & 70.88 
& 64.95 & 70.73 \\

& X-JEPA  
& \underline{79.12} & \underline{85.19} 
& \underline{67.34} & \underline{70.78} 
& \underline{70.11} & \underline{74.41} 
& \underline{67.87} & \underline{73.41} \\

& \method 
& \textbf{86.94} & \textbf{90.94} 
& \textbf{71.88} & \textbf{76.21} 
& \textbf{72.62} & \textbf{78.11} 
& \textbf{73.55} & \textbf{78.41} \\
\midrule
Dataset & Method
& \multicolumn{2}{c}{PAN$\rightarrow$PAN}
& \multicolumn{2}{c}{MS$\rightarrow$MS}
& \multicolumn{2}{c}{PAN$\rightarrow$MS}
& \multicolumn{2}{c}{MS$\rightarrow$PAN} \\
\cmidrule(lr){3-4}\cmidrule(lr){5-6}\cmidrule(lr){7-8}\cmidrule(lr){9-10}
& & mAP & P@5 & mAP & P@5 & mAP & P@5 & mAP & P@5 \\
\midrule
\multirow{3}{*}{DSRSID}
& REJEPA  
& 62.42 & 66.18 
& 65.76 & 69.35 
& 62.21 & 67.62 
& 65.84 & 69.37 \\

& X-JEPA  
& \underline{64.65} & \underline{68.42}
& \underline{72.13} & \underline{76.08}
& \underline{66.36} & \underline{70.74} 
& \underline{68.82} & \underline{72.91} \\

& \method 
& \textbf{69.82} & \textbf{77.81} 
& \textbf{73.10} & \textbf{79.13} 
& \textbf{72.15} & \textbf{77.27} 
& \textbf{71.24} & \textbf{78.43} \\
\bottomrule
\end{tabular}
}
\caption{Retrieval results on CBRSIR\_VS and DSRSID under single-label relevance. CBRSIR\_VS uses RGB optical and SAR modalities, while DSRSID uses panchromatic (PAN) and multispectral (MS) modalities. We report global mAP and P@5. Best and second-best results are shown in bold and underlined, respectively.}
\label{tab:cbrsir_dsrsid_results}
\end{table*}

On CBRSIR\_VS, X-JEPA consistently improves over REJEPA across both same-modal and cross-modal retrieval directions. However, \method achieves the best performance in all four directions. For same-modal retrieval, \method improves RGB$\rightarrow$RGB to $86.94$ mAP and SAR$\rightarrow$SAR to $71.88$ mAP. For cross-modal retrieval, \method reaches $72.62$ mAP for RGB$\rightarrow$SAR and $73.55$ mAP for SAR$\rightarrow$RGB, outperforming X-JEPA by $2.51$ and $5.68$ mAP points, respectively. The P@5 scores show the same trend, with \method achieving $78.11$ and $78.41$ in the two cross-modal directions.

On DSRSID, \method also achieves the best performance across all retrieval directions. Compared with X-JEPA, \method improves PAN$\rightarrow$PAN from $64.65$ to $69.82$ mAP and from $68.42$ to $77.81$ P@5. For MS$\rightarrow$MS retrieval, the improvement is more moderate, increasing from $72.13$ to $73.10$ mAP and from $76.08$ to $79.13$ P@5. In cross-modal retrieval, \method improves PAN$\rightarrow$MS from $66.36$ to $72.15$ mAP and MS$\rightarrow$PAN from $68.82$ to $71.24$ mAP. The corresponding P@5 scores also increase from $70.74$ to $77.27$ and from $72.91$ to $78.43$, respectively.

Overall, these results show that \method consistently improves over REJEPA and X-JEPA on both CBRSIR\_VS and DSRSID. The gains are especially clear in cross-modal retrieval and remain positive in same-modal retrieval, indicating that the proposed predictive learning, decoupled retrieval heads, and SIGReg regularization improve retrieval across different heterogeneous modality pairs.

\subsection{Ablation Study}

We conduct ablation studies on BEN-14K to analyze the contribution of the main architectural and training components of \method. Unless otherwise stated, all variants are trained and evaluated under the same protocol as the full model, and F1@5 is reported for the four retrieval directions. In the main paper, we report sensitivity to the mask ratio and predictor depth; additional component-level ablations covering loss design, predictor routing, predictor sharing strategies, retrieval-loss combinations, and shared-versus-separate trunk configurations are provided in the supplementary material.

\begin{table*}[ht]
\centering
\scriptsize
\setlength{\tabcolsep}{3.2pt}
\renewcommand{\arraystretch}{1.08}
\resizebox{\textwidth}{!}{%
\begin{tabular}{lccc|cccccc}
\toprule
\multirow{2}{*}{Direction}
& \multicolumn{3}{c|}{Mask ratio}
& \multicolumn{6}{c}{Predictor depth} \\
\cmidrule(lr){2-4}\cmidrule(lr){5-10}
& 0.25 & \textbf{0.50} & 0.75
& 2 & 4 & \textbf{6} & 8 & 10 & 12 \\
\midrule
S1$\rightarrow$S1
& 70.12 & \textbf{75.11} & \underline{70.48}
& 70.11 & 70.72 & \textbf{75.11} & \underline{71.93} & 70.56 & 70.44 \\

S2$\rightarrow$S2
& 76.69 & \textbf{82.87} & \underline{77.87}
& 74.81 & 77.77 & \textbf{82.87} & 79.04 & \underline{80.03} & 75.74 \\

S1$\rightarrow$S2
& 70.87 & \textbf{75.82} & \underline{72.03}
& 71.54 & 71.52 & \textbf{75.82} & \underline{74.49} & 71.01 & 71.74 \\

S2$\rightarrow$S1
& 70.89 & \textbf{75.40} & \underline{72.10}
& 71.64 & \underline{71.91} & \textbf{75.40} & 71.16 & 71.63 & 71.79 \\
\bottomrule
\end{tabular}
}
\caption{Sensitivity analysis on BEN-14K. The default configuration uses mask ratio $0.50$ and predictor depth $6$. Each entry reports F1@5. Best results are in bold and second-best results are underlined.}
\label{tab:ablation_sensitivity}
\end{table*}

\paragraph{Sensitivity to masking and predictor depth.} Table~\ref{tab:ablation_sensitivity} reports the sensitivity of \method to the mask ratio and predictor depth. The mask ratio controls the difficulty of the JEPA-style latent prediction task. A moderate mask ratio of $0.50$ gives the best performance across all retrieval directions, while both lower and higher masking reduce retrieval quality. Predictor depth also affects performance: a depth of $6$ provides the best overall results, suggesting that sufficient predictor capacity is important for masked latent target estimation, but deeper predictors do not further improve retrieval. 


\subsection{Qualitative Retrieval Analysis}

\begin{figure}
    \centering
    \includegraphics[width=\linewidth]{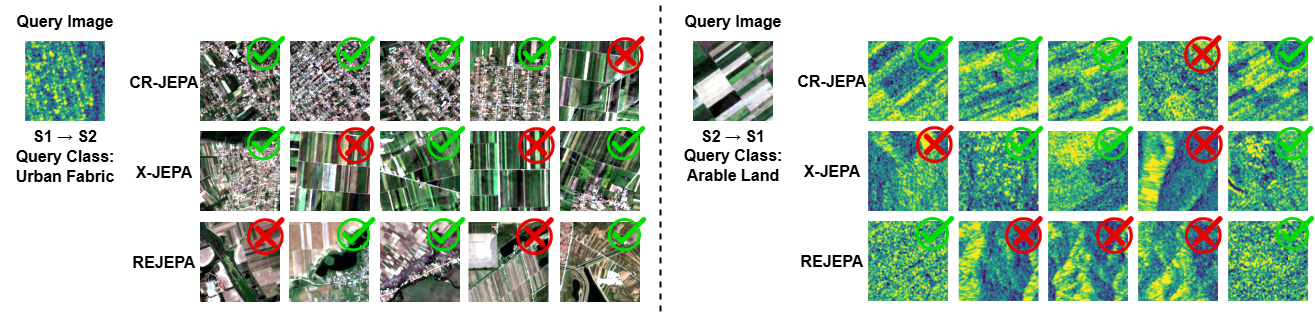}
    \caption{Qualitative cross-modal retrieval examples on BEN-14K. Each block shows a query image and its top-5 retrieved results under different methods. Correct retrievals are marked in green and incorrect ones in red.}
    \label{fig:ben14k_qr}
\end{figure}

The quantitative results are complemented with qualitative retrieval examples. Figure~\ref{fig:ben14k_qr} presents representative query and top-K retrieval grids from BEN-14K, illustrating cross-modal retrieval performance under the same evaluation protocol as Table~\ref{tab:main_results_ben14k}. Each row corresponds to one retrieval direction, and retrieved samples are marked to indicate correct and incorrect matches. The examples demonstrate that \method retrieves semantically consistent scenes across modalities, even when substantial appearance differences exist between S1 and S2 imagery. Failure cases are also included to highlight typical error patterns, such as semantic overlap between visually similar land-cover categories or ambiguity caused by structural similarity across classes. For completeness, additional qualitative retrieval results on CBRSIR\_VS and DSRSID are provided in the supplementary material.

\subsection{Discussion}

The results show that \method is particularly effective for cross-modal retrieval while maintaining strong same-modal performance. On BEN-14K, the large gains in S1$\rightarrow$S2 and S2$\rightarrow$S1 retrieval indicate that latent predictive learning, SIGReg-based embedding regularization, and retrieval-space specialization improve semantic alignment across heterogeneous sensing modalities. The same-modal improvements further suggest that the proposed architecture preserves within-modality neighbourhood structure while learning cross-modal correspondence.

Across the three benchmarks, \method remains effective under different modality gaps: Sentinel-1/Sentinel-2 in BEN-14K, RGB optical/SAR in CBRSIR\_VS, and panchromatic /multispectral imagery in DSRSID. This consistency suggests that \method is not limited to a specific sensor pair, but provides a general framework for heterogeneous dual-modality remote sensing retrieval.

\section{Conclusion and Future Work}

We presented \method, a JEPA-based framework for heterogeneous dual-modality remote sensing image retrieval. The proposed architecture combines modality-specific stems, a shared semantic transformer trunk, same-modal and cross-modal latent predictive learning, SIGReg-based embedding regularization, and decoupled retrieval heads. This design separates low-level sensor adaptation, shared semantic reasoning, and retrieval-space specialization for same-modal and cross-modal retrieval.

Experiments on BEN-14K, CBRSIR\_VS, and DSRSID show consistent improvements over recent self-supervised and JEPA-based retrieval baselines, with especially strong gains in cross-modal directions. On BEN-14K, \method improves S1$\rightarrow$S2 and S2$\rightarrow$S1 retrieval over X-JEPA by 14.59 and 11.67 F1@5 points, respectively, while using 117.93M parameters compared with 172.86M for X-JEPA. On CBRSIR\_VS and DSRSID, \method also achieves the best performance across all same-modal and cross-modal retrieval directions, confirming its effectiveness beyond Sentinel-1/Sentinel-2 retrieval. Overall, the results support \method as a general retrieval framework for heterogeneous dual-modality remote sensing data.

Future work will explore scaling \method toward remote sensing foundation-model pretraining across more sensor combinations, stronger retrieval heads for multi-sensor and multi-resolution Earth observation data, and extension to partially paired or unpaired archives.

\bibliography{references}

\clearpage
\setcounter{section}{0}
\setcounter{subsection}{0}
\setcounter{table}{0}
\setcounter{figure}{0}
\setcounter{equation}{0}
\renewcommand{\thesection}{S\arabic{section}}
\renewcommand{\thesubsection}{S\arabic{section}.\arabic{subsection}}
\renewcommand{\thetable}{S\arabic{table}}
\renewcommand{\thefigure}{S\arabic{figure}}
\renewcommand{\theequation}{S\arabic{equation}}

{\centering
\Large\bfseries Supplementary Material for CR-JEPA: Cross-Modal Joint-Embedding Predictive Learning for Remote Sensing Image Retrieval\par
}
\vspace{1em}

\section{Supplementary Overview}

This supplementary material provides additional details supporting the main paper. Section~\ref{suppl:datasets} describes the datasets and modality setups. Section~\ref{suppl:metrics} includes the full metric definitions. Section~\ref{suppl:qualitative} offers more qualitative retrieval examples. Section~\ref{suppl:ablations} presents extended ablation studies on BEN-14K. Section~\ref{suppl:dual_encoder_variant} outlines a dual-encoder variant that analyzes the benefit of the shared semantic trunk.

\section{Dataset Details}
\label{suppl:datasets}

We evaluate \method on three dual-modality remote sensing retrieval benchmarks: BEN-14K, CBRSIR\_VS, and DSRSID. These datasets involve Sentinel-1/Sentinel-2, RGB optical/SAR, and panchromatic/multispectral retrieval settings. Table~\ref{tab:suppl_datasets} summarizes the dataset properties used in our experiments.

\textbf{BEN-14K.}
BEN-14K is derived from BigEarthNet-MM and contains 14,832 paired Sentinel-1 SAR and Sentinel-2 multispectral image pairs acquired over Serbia during summer~\cite{sumbul2021bigearthnetmm,bigearthnet}. Each sample includes one Sentinel-1 image, one Sentinel-2 image, and a multi-label land-cover annotation vector. We use BEN-14K to assess same-modal and cross-modal retrieval under multi-label relevance, following recent sensor-agnostic retrieval protocols~\cite{hackstein2025csmae,choudhury2026xjepa}.

\textbf{CBRSIR\_VS.}
CBRSIR\_VS is an optical/SAR dual-modality retrieval dataset introduced for cross-modal remote sensing retrieval and hashing~\cite{sun2022msesph}. It contains 26,901 paired RGB optical and SAR images from 10 semantic classes. The optical images are $256\times256$ very-high-resolution RGB images with 1\,m spatial resolution, while the SAR images are Sentinel-1 images of size $64\times64$ with 10\,m spatial resolution.

\textbf{DSRSID.}
DSRSID is a dual-source dataset for cross-source remote sensing image retrieval~\cite{li2018sidhcnn}. It contains 80,000 paired panchromatic/multispectral samples acquired by the Gaofen-1 optical satellite. The panchromatic images are one-channel $256\times256$ images with 2\,m spatial resolution, while the multispectral images have 4 channels and size $64\times64$ with 8\,m spatial resolution. DSRSID contains eight classes: aquafarm, cloud, forest, high building, low building, farm land, river, and water.

\begin{table}[ht]
\centering
\small
\setlength{\tabcolsep}{4pt}
\renewcommand{\arraystretch}{1.1}
\begin{tabular}{lcccc}
\toprule
Dataset & Modality pair & Classes & Pairs & Relevance type \\
\midrule
BEN-14K~\cite{sumbul2021bigearthnetmm} & Sentinel-1 / Sentinel-2 & 19 & 14,832 & Multi-label overlap \\
CBRSIR\_VS~\cite{sun2022msesph} & RGB optical / SAR & 10 & 26,901 & Single-label equality \\
DSRSID~\cite{li2018sidhcnn} & PAN / Multispectral & 8 & 80,000 & Single-label equality \\
\bottomrule
\end{tabular}
\caption{Summary of the datasets used for evaluating \method.}
\label{tab:suppl_datasets}
\end{table}

\section{Evaluation Metric Definitions}
\label{suppl:metrics}

We evaluate retrieval in four directions:
\begin{equation}
a\rightarrow a,\qquad b\rightarrow b,\qquad a\rightarrow b,\qquad b\rightarrow a .
\end{equation}
The first two denote same-modal retrieval, while the last two denote cross-modal retrieval. 
Trivial self-matches are excluded in same-modal retrieval when the query and gallery contain the same indexed sample.

\textbf{Multi-label relevance on BEN-14K.}
For BEN-14K, relevance is defined by multi-label overlap. Let $\mathcal{Y}_q$ and $\mathcal{Y}_r$ denote the label sets of a query image and a retrieved image, respectively. For each retrieved item, we compute label-overlap precision, recall, and F1 as
\begin{equation}
P(q,r)=\frac{|\mathcal{Y}_q\cap \mathcal{Y}_r|}{|\mathcal{Y}_r|}, \qquad
R(q,r)=\frac{|\mathcal{Y}_q\cap \mathcal{Y}_r|}{|\mathcal{Y}_q|},
\end{equation}
\begin{equation}
F_1(q,r)=\frac{2P(q,r)R(q,r)}{P(q,r)+R(q,r)+\epsilon}.
\end{equation}
The final F1@K is obtained by averaging the item-level F1 scores over the top-$K$ retrieved images and then over all queries. We report F1@5 for BEN-14K, following recent remote sensing retrieval protocols~\cite{hackstein2025csmae,choudhury2026xjepa}.

\textbf{Single-label relevance on CBRSIR\_VS and DSRSID.}
For CBRSIR\_VS and DSRSID, relevance is defined by single-label class equality. A retrieved image is considered relevant if it belongs to the same semantic class as the query. For a query $q$, let $\mathrm{rel}_q(k)=1$ if the gallery image retrieved at rank $k$ is relevant, and $0$ otherwise. Precision at rank $K$ is
\begin{equation}
P@K(q)=\frac{1}{K}\sum_{k=1}^{K}\mathrm{rel}_q(k).
\end{equation}
We use $K=5$ and report P@5. We also report mean average precision (mAP) over the full gallery. For query $q$, precision at rank $k$ is
\begin{equation}
P_q(k)=\frac{1}{k}\sum_{i=1}^{k}\mathrm{rel}_q(i).
\end{equation}
The average precision for query $q$ is
\begin{equation}
AP(q)=\frac{1}{R_q}\sum_{k=1}^{|G|}P_q(k)\,\mathrm{rel}_q(k),
\end{equation}
where $R_q$ is the number of relevant gallery samples for query $q$, and $|G|$ is the gallery size. The final mAP is
\begin{equation}
mAP=\frac{1}{|Q|}\sum_{q\in Q}AP(q).
\end{equation}

\section{Additional Qualitative Retrieval Results}
\label{suppl:qualitative}

\begin{figure}[ht]
    \centering
    \includegraphics[width=\linewidth]{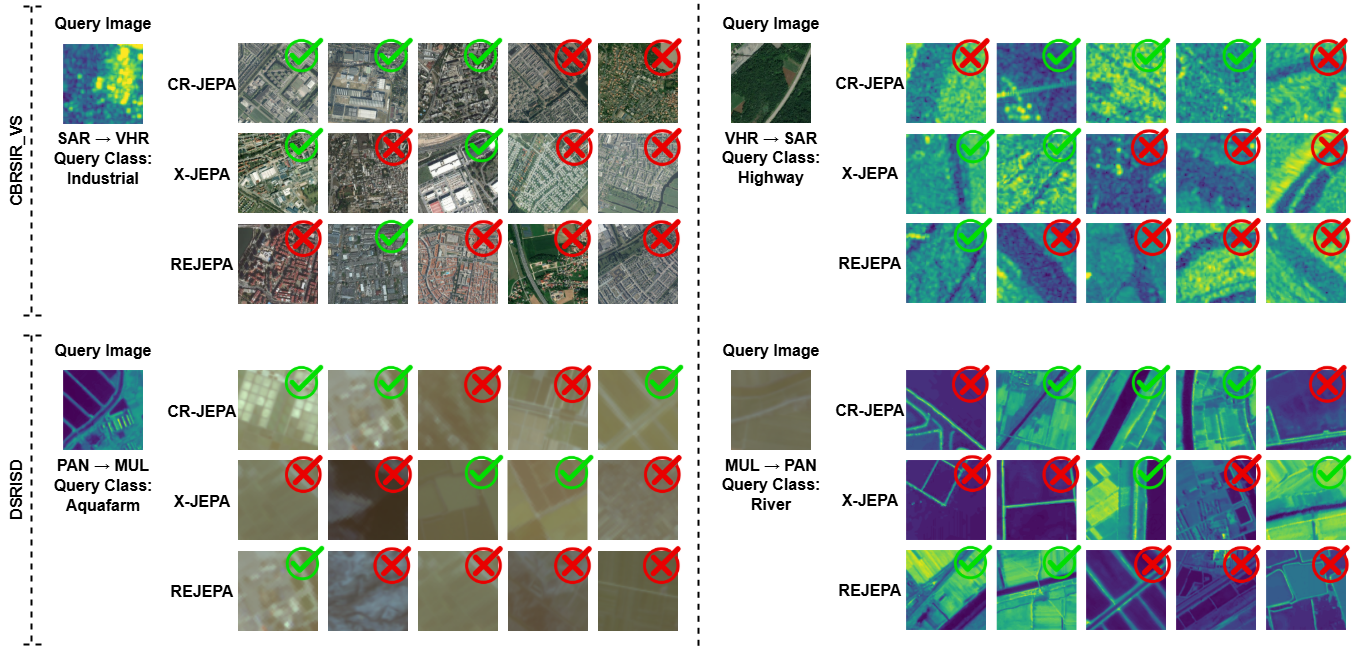}
    \caption{Additional qualitative retrieval examples on CBRSIR\_VS and DSRSID. These examples illustrate cross-modal retrieval across RGB optical/SAR and PAN/multispectral modality pairs.}
    \label{fig:suppl_cbrsir_dsrsid_qr}
\end{figure}

Figure~\ref{fig:suppl_cbrsir_dsrsid_qr} offers more qualitative retrieval examples on CBRSIR\_VS and DSRSID. These examples support the BEN-14K qualitative findings in the main paper and demonstrate that \method retrieves semantically relevant samples across RGB optical/SAR and panchromatic/multispectral modality pairs.

\section{Additional Ablation Studies}
\label{suppl:ablations}

We provide detailed ablation studies on BEN-14K to support the main paper. All entries show F1@5 using the same four-direction retrieval protocol as in the main experiments. Unless mentioned otherwise, the final configuration includes all predictive routes, the shared semantic trunk, the cross-modal retrieval head, the unified retrieval head, and SIGReg regularization.

\subsection{Loss Components and Predictor Routing}

Table~\ref{tab:suppl_joint_ablation}(a) evaluates the contribution of predictive learning, cross-modal retrieval loss, unified retrieval loss, and SIGReg. $\mathcal{L}_{\mathrm{pred}}$ denotes the JEPA-style predictive objective, $\mathcal{L}_{\mathrm{cross}}$ denotes the cross-modal retrieval loss, $\mathcal{L}_{\mathrm{uni}}$ denotes the unified retrieval loss, and $\mathcal{L}_{\mathrm{sigreg}}$ denotes the SIGReg embedding regularizer. The full objective achieves the best average performance across the four retrieval directions, indicating that the losses provide complementary supervision. Some partial combinations are competitive in individual directions, but they are less stable across all retrieval modes.

\begin{table*}[ht]
\centering
\scriptsize
\setlength{\tabcolsep}{3pt}
\renewcommand{\arraystretch}{1.08}
\resizebox{\textwidth}{!}{%
\begin{tabular}{cccc|ccccc}
\toprule
\multicolumn{9}{c}{\textbf{(a) Loss Component Ablation}} \\
\midrule
$\mathcal{L}_{\mathrm{pred}}$ & $\mathcal{L}_{\mathrm{cross}}$ & $\mathcal{L}_{\mathrm{uni}}$ & $\mathcal{L}_{\mathrm{sigreg}}$
& S1$\rightarrow$S1 & S2$\rightarrow$S2 & S1$\rightarrow$S2 & S2$\rightarrow$S1 & Avg. \\
\midrule
\checkmark & \checkmark &  &  & 73.48 & 78.54 & 47.14 & 45.99 & 61.29 \\
\checkmark &  & \checkmark &  & 74.44 & 77.68 & 75.38 & 74.79 & 75.57 \\
\checkmark &  &  & \checkmark & 49.04 & 65.56 & 36.89 & 36.51 & 47.00 \\
\checkmark & \checkmark & \checkmark &  & 74.53 & 77.89 & 74.32 & 75.01 & 75.44 \\
\checkmark & \checkmark &  & \checkmark & 73.79 & 78.42 & 49.59 & 48.13 & 62.48 \\
\checkmark &  & \checkmark & \checkmark & 74.01 & 78.60 & 75.24 & \textbf{75.55} & 75.85 \\
\checkmark & \checkmark & \checkmark & \checkmark & \textbf{75.11} & \textbf{82.87} & \textbf{75.82} & 75.40 & \textbf{77.30} \\
\midrule
\multicolumn{9}{c}{\textbf{(b) Predictor Routing Ablation}} \\
\midrule
$\mathcal{L}_{S11}$ & $\mathcal{L}_{S22}$ & $\mathcal{L}_{S12}$ & $\mathcal{L}_{S21}$
& S1$\rightarrow$S1 & S2$\rightarrow$S2 & S1$\rightarrow$S2 & S2$\rightarrow$S1 & Avg. \\
\midrule
\checkmark & \checkmark &  &  & 75.04 & 79.62 & 75.44 & 74.77 & 76.23 \\
 &  & \checkmark & \checkmark & 75.10 & 81.03 & \textbf{75.97} & 75.35 & 76.86 \\
\checkmark &  & \checkmark &  & \textbf{75.67} & 80.92 & 76.13 & 75.18 & 76.98 \\
 & \checkmark &  & \checkmark & 75.34 & 79.82 & 75.94 & 75.36 & 76.62 \\
\checkmark & \checkmark & \checkmark & \checkmark & 75.11 & \textbf{82.87} & 75.82 & \textbf{75.40} & \textbf{77.30} \\
\bottomrule
\end{tabular}
}
\caption{Joint ablation study on BEN-14K. (a) evaluates objective components, while (b) evaluates predictor routing across same-modal and cross-modal prediction tasks. The final configuration achieves the best average F1@5.}
\label{tab:suppl_joint_ablation}
\end{table*}

Table~\ref{tab:suppl_joint_ablation}(b) studies the effect of predictor routing. $\mathcal{L}_{S11}$ and $\mathcal{L}_{S22}$ denote same-modal predictive losses, while $\mathcal{L}_{S12}$ and $\mathcal{L}_{S21}$ denote cross-modal predictive losses. The results show that jointly using all same-modal and cross-modal routes achieves the best average performance, supporting the bidirectional predictive design used in the final model.

\begin{table*}[!t]
\centering
\small
\setlength{\tabcolsep}{4pt}
\renewcommand{\arraystretch}{1.1}
\resizebox{\textwidth}{!}{%
\begin{tabular}{lcccc}
\toprule
Predictor sharing strategy & S1$\rightarrow$S1 & S2$\rightarrow$S2 & S1$\rightarrow$S2 & S2$\rightarrow$S1 \\
\midrule
Fully independent predictors & 74.97 & 78.42 & 75.72 & 74.63 \\
Shared same-modal + shared cross-modal & 75.04 & 78.80 & 75.67 & 75.40 \\
Single shared predictor for all tasks & 74.70 & 78.38 & 75.56 & 75.42 \\
Shared same-modal + independent cross-modal & 74.70 & 78.80 & 75.67 & 75.40 \\
Modality-specific same + shared cross (final) & \textbf{75.11} & \textbf{82.87} & \textbf{75.82} & 75.40 \\
\bottomrule
\end{tabular}
}
\caption{Ablation study of predictor sharing strategies on BEN-14K. Each entry reports F1@5.}
\label{tab:suppl_predictor_sharing}
\end{table*}

\subsection{Predictor Sharing Strategy}

Table~\ref{tab:suppl_predictor_sharing} evaluates how predictor sharing affects retrieval. The final design uses modality-specific predictors for same-modal prediction and a shared predictor for cross-modal prediction. This configuration preserves modality-specific forecasting capacity while enforcing a common cross-modal predictive structure.

\subsection{Retrieval-Loss Ablation}
\label{suppl:retrieval_loss_ablation}
Table~\ref{tab:suppl_retrieval_loss_ablation} reports retrieval-loss and head-design ablations using an early three-head variant of \method. This variant used modality-specific same-modal retrieval heads with a same-modal retrieval regularization loss $\mathcal{L}_{\mathrm{same}}$, a shared cross-modal retrieval head with cross-modal retrieval loss $\mathcal{L}_{\mathrm{cross}}$, and a unified retrieval head with unified retrieval loss $\mathcal{L}_{\mathrm{uni}}$. In the final model, we remove the explicit same-modal retrieval heads and the associated loss $\mathcal{L}_{\mathrm{same}}$, retaining only $\mathcal{L}_{\mathrm{cross}}$ and $\mathcal{L}_{\mathrm{uni}}$.

For modality $m\in\{a,b\}$, the same-modal retrieval embeddings in the early three-head variant are
\begin{equation}
r_{\mathrm{same}}^{(m)}, e_{\mathrm{same}}^{(m)}
=
\phi_{\mathrm{same}}^{(m)}(\bar{z}^{(m)}).
\end{equation}
The same-modal retrieval regularization is
\begin{equation}
\mathcal{L}_{\mathrm{same}}
=
\mathrm{NCE}(e_{\mathrm{same}}^{(a)},\mathrm{sg}(e_{\mathrm{same}}^{(a)}))
+
\mathrm{NCE}(e_{\mathrm{same}}^{(b)},\mathrm{sg}(e_{\mathrm{same}}^{(b)})),
\end{equation}
where $\mathrm{sg}(\cdot)$ denotes stop-gradient. The final model removes this term and uses
\begin{equation}
\mathcal{L}_{\mathrm{retr}}
=
\lambda_{\mathrm{cross}}\mathcal{L}_{\mathrm{cross}}
+
\lambda_{\mathrm{uni}}\mathcal{L}_{\mathrm{uni}}.
\end{equation}

\begin{table*}[ht]
\centering
\small
\setlength{\tabcolsep}{4pt}
\renewcommand{\arraystretch}{1.1}
\begin{tabular}{lcccccccc}
\toprule
Variant 
& $\mathcal{L}_{\mathrm{same}}$
& $\mathcal{L}_{\mathrm{cross}}$
& $\mathcal{L}_{\mathrm{uni}}$
& S1$\rightarrow$S1
& S2$\rightarrow$S2
& S1$\rightarrow$S2
& S2$\rightarrow$S1
& Avg. \\
\midrule
Same + Cross & \checkmark & \checkmark & -- & 67.72 & 71.43 & 24.13 & 35.46 & 49.68 \\
Same + Unified & \checkmark & -- & \checkmark & 68.06 & 73.54 & 71.63 & 71.18 & 71.10 \\
Same + Cross + Unified & \checkmark & \checkmark & \checkmark & 74.02 & 76.11 & 75.03 & 74.92 & 75.02 \\
Final \method & -- & \checkmark & \checkmark & 75.11 & 82.87 & 75.82 & 75.40 & 77.30 \\
\bottomrule
\end{tabular}
\caption{Retrieval-loss and head-design ablation on BEN-14K. Each entry reports F1@5. The final model removes the explicit same-modal retrieval heads and the associated same-modal retrieval loss, using the cross-modal retrieval loss together with the unified retrieval loss.}
\label{tab:suppl_retrieval_loss_ablation}
\end{table*}

These ablations verify whether explicit same-modal retrieval regularization is necessary. The results show that adding $\mathcal{L}_{\mathrm{same}}$ does not improve the final configuration. Therefore, the final model simplifies the retrieval objective by removing $\mathcal{L}_{\mathrm{same}}$, while preserving strong same-modal performance and improving cross-modal retrieval.

\subsection{Shared versus Separate Trunks}

Table~\ref{tab:suppl_shared_vs_separate} compares the final shared-trunk design with the dual-encoder CR-JEPA variant, where the two modalities are processed by separate transformer trunks. The shared-trunk model uses modality-specific stems for low-level sensor adaptation, followed by a common semantic trunk for both modalities. In contrast, the dual-encoder variant maintains independent modality-specific encoders throughout the network.

\begin{table}[ht]
\centering
\small
\setlength{\tabcolsep}{5pt}
\renewcommand{\arraystretch}{1.1}
\begin{tabular}{lccccc}
\toprule
Setting & Params & S1$\rightarrow$S1 & S2$\rightarrow$S2 & S1$\rightarrow$S2 & S2$\rightarrow$S1 \\
\midrule
Shared trunk & \textbf{117.93M} & \textbf{75.11} & \textbf{82.87} & \textbf{75.82} & \textbf{75.40} \\
Separate trunks & 154.77M & 70.34 & 75.80 & 67.34 & 71.68 \\
\bottomrule
\end{tabular}
\caption{Comparison between shared-trunk and separate-trunk architectures on BEN-14K. The separate-trunk setting corresponds to the dual-encoder CR-JEPA variant. Each entry reports F1@5.}
\label{tab:suppl_shared_vs_separate}
\end{table}

The shared-trunk design reduces trainable parameters and improves retrieval performance in all four directions. This suggests that sharing the higher-level transformer trunk encourages modality-common semantic reasoning, while avoiding over-specialization of two independent encoders. These results support the final design choice of using modality-specific stems followed by a shared semantic trunk.

\section{Dual-Encoder CR-JEPA Variant}
\label{suppl:dual_encoder_variant}
The separate-trunk configuration in Table~\ref{tab:suppl_shared_vs_separate} is implemented as a dual-encoder CR-JEPA variant. Unlike the final architecture, which uses modality-specific stems followed by a shared transformer trunk, this variant uses two independent ViT encoders, $E_a$ and $E_b$, for the two modalities. Each encoder has its own patch embedding, positional embeddings, transformer blocks, and normalization layer. Thus, the two modalities are processed in separate feature spaces until retrieval and alignment losses are applied. This design provides a direct comparison for evaluating whether independent modality-specific encoders are sufficient or whether a shared semantic trunk better supports heterogeneous remote sensing retrieval.

For an input pair $(x^{(a)},x^{(b)})$, the encoders produce visible-context and masked-target latent tokens for each modality. The same-modal predictors estimate masked tokens within each modality, while a shared cross-modal predictor estimates masked target tokens of one modality from the visible context of the other modality. The predictive loss follows the same four-path objective used in the main model:
\begin{equation}
\mathcal{L}_{\mathrm{pred}}
=
\lambda_{aa}\mathcal{L}_{aa}
+
\lambda_{bb}\mathcal{L}_{bb}
+
\lambda_{ab}\mathcal{L}_{ab}
+
\lambda_{ba}\mathcal{L}_{ba}.
\end{equation}

For retrieval, the dual-encoder variant uses a single shared global retrieval head $\phi_{\mathrm{retr}}$ applied to the pooled tokens from both encoders:
\begin{equation}
r^{(m)}, e^{(m)} = \phi_{\mathrm{retr}}(z^{(m)}), \qquad m\in\{a,b\}.
\end{equation}
The retrieval objective combines cross-modal InfoNCE and direct cosine alignment:
\begin{equation}
\mathcal{L}_{\mathrm{retr}}
=
\mathrm{NCE}(e^{(a)},e^{(b)})
+
\frac{1}{B}\sum_{i=1}^{B}
\left(1-\cos(e_i^{(a)},e_i^{(b)})\right).
\end{equation}

In addition, this variant includes two SIGReg-based regularization terms. First, SIGReg is applied to the raw retrieval projections from both modalities to improve the conditioning of the embeddings. Second, a paired cross-modal SIGReg term is applied after fusing the raw projections from the two modalities. In our implementation, the paired SIGReg term is computed by applying SIGReg to the fused paired raw projections, obtained by averaging the raw projections from the two modalities. The total objective is
\begin{equation}
\mathcal{L}
=
\mathcal{L}_{\mathrm{pred}}
+
\lambda_{\mathrm{retr}}\mathcal{L}_{\mathrm{retr}}
+
\lambda_{\mathrm{sigreg}}\mathcal{L}_{\mathrm{sigreg}}
+
\lambda_{\mathrm{pair}}\mathcal{L}_{\mathrm{pair\text{-}sigreg}} .
\end{equation}

This dual-encoder design provides a useful comparison against the final \method. It tests whether independent modality encoders with retrieval alignment and SIGReg-based regularization are sufficient, or whether modality-specific stems followed by a shared semantic trunk provide better semantic coupling for heterogeneous remote sensing retrieval.

\end{document}